\documentclass[fleqn,10pt]{wlscirep}
\usepackage[utf8]{inputenc}
\usepackage[T1]{fontenc}
\usepackage{lineno}
\usepackage[acronym]{glossaries}
\usepackage{listings}
\usepackage{xcolor}

\makeglossaries

\newacronym{ems}{EMS}{Energy management system}
\newacronym{dsm}{DSM}{Demand-side management}
\newacronym{dr}{DR}{Demand response}
\newacronym{rdf}{RDF}{Resource Description Framework}
\newacronym{ict}{ICT}{Information and communications technologies}
\newacronym{iot}{IoT}{Internet of things}
\newacronym{nilm}{NILM}{Non-intrusive load monitoring}
\newacronym{ilm}{ILM}{Intrusive load monitoring}
\newacronym{cnn}{CNN}{Convolutional neural network}
\newacronym{gdp}{GDP}{Gross domestic product}
\newacronym{r2rml}{R2RML}{Relational database to \gls{rdf} mapping language}
\newacronym{ml}{ML}{Machine learning}
\newacronym{kg}{KG}{Knowledge graph}
\newcommand{\acrfullnohyper}[1]{\glsentrylong{#1} (\glsentryshort{#1})}
\newcommand{\acrshortnohyper}[1]{\glsentryshort{#1}}

\renewcommand{\gls}[1]{%
  \ifglsused{#1}{%
    \acrshortnohyper{#1}%
  }{%
    \glsunset{#1}
    \acrfullnohyper{#1}%
  }%
}

\title{Towards Data-Driven Electricity Management: Multi-Region Harmonized Data and Knowledge Graph}

\author[1,*]{Vid Han\v{z}el}
\author[1]{Bla\v{z} Bertalani\v{c}}
\author[1]{Carolina Fortuna}
\affil[1]{Jozef Stefan Institute,  Ljubljana, 1000, Slovenia}

\affil[*]{corresponding author(s): Vid Hanžel (vid.hanzel@gmail.com)}

\begin{abstract}
Due to growing population and technological advances, global electricity consumption, and consequently also CO\textsubscript{2} emissions are increasing. The residential sector makes up 25\% of global electricity consumption and has great potential to increase efficiency and reduce CO\textsubscript{2} footprint without sacrificing comfort. 
However, a lack of uniform consumption data at the household level spanning multiple regions hinders large-scale studies and robust multi-region model development.
This paper introduces a multi-region dataset compiled from publicly available sources and presented in a uniform format. This data enables machine learning tasks such as disaggregation, demand forecasting, appliance ON/OFF classification, etc. Furthermore, we develop an RDF knowledge graph that characterizes the electricity consumption of the households and contextualizes it with household related properties enabling semantic queries and interoperability with other open knowledge bases like Wikidata and DBpedia. This structured data can be utilized to inform various stakeholders towards data-driven policy and business development.

\end{abstract}
\begin{document}

\flushbottom
\maketitle

\thispagestyle{empty}

\section*{Background \& Summary}

Increased energy consumption that comes with population growth and technological advances raises environmental concerns, therefore many public and private institutions are committing to emission cuts and increased energy efficiency solutions. For instance, the  European Union has committed to improving energy efficiency by at least 32.5\% by 2030~\footnote{\url{https://ec.europa.eu/clima/eu-action/climate-strategies-targets/2030-climate-energy-framework\_en}}. Efforts in meeting such targets are underway and involve i) increasing efficiency of existing technologies, such as engines, home appliances and electricity grids; ii) moving towards renewable sources and iii) developing incentives for various stakeholders. While success stories are being reported, such as the case of Denmark where in 2017, 70.6\% of the total electricity production was generated by renewable energies, including wind (67.4\%), biomass (25.9\%), solar (3.4\%), biogas
(3.1\%), and hydro (0.1\%)~\cite{TROTTA2020115399}, half of the world's electricity is still generated from natural gas and coal~\cite{IEA2027_technology}. Nevertheless, to fulfill projections such as having renewables contribute to 38\% of global electricity generation~\cite{IEA2027_technology} by 2027, suitable policies and regulations to incentivize all stakeholders towards the desired goal have to be developed. Focusing on achieving the overall sustainability goals, we identify three groups of stakeholders, namely \textit{Governments and regulatory bodies}, \textit{Distribution system operators} and \textit{Household residents} and analyze their role in achieving the targets and the challenges they are confronted with.

\textbf{Government and Regulatory Bodies:} The role of such stakeholders is to design good policies and regulations to achieve an overall increase in the share of renewables in household consumption. In this endeavor, social justice aspects also have to be considered as it has been reported that the capacity of energy users to shift their energy-using practices in time or space, also referred to as flexibility capital, makes some householders more capable of being flexible, than others~\cite{FJELLSA202198}. Given this, more comprehensive data-driven studies to understand different residential electricity consumption behaviors~~\cite{anvari2022data}, effects of moving or rescheduling flexible load in view of peak shaving/clipping, etc. are required as most of the existing studies are limited by the measured metrics, sample size, period, and geographical scale~\cite{TROTTA2020115399, Yu2024, Cong2022}. For instance, the energy poverty index is typically calculated by the share of disposable income being spent on energy~\cite{SOVACOOL2015361}. However, Cong~\cite{Cong2022} showed that, when more information is available, namely the outside temperature and exact timings when a household turns on the heating, more accurate metrics can be developed better informing subsidizing policies for such households. Furthermore, in China the government aims to implement a carbon peak initiative, reaching peak carbon emissions by 2030 and achieving carbon neutrality by 2060. The realization of this goal is hindered by the lack of comprehensive CO\textsubscript{2} emission data~\cite{Yu2024}. 

\textbf{Distribution System Operators:} These are companies that aim to deliver electricity with the highest possible efficiency by managing the electric grid effectively. Based on policies, regulations, and market constraints, they are undergoing deep technological and business model transformations. Their distribution grid is being upgraded with electrical and \gls{ict} components becoming a so-called smart grid that allows bidirectional transmission of power and data, creating an automated power grid~\cite{FangSG}. With smart grids, every actor, including individual households equipped with solar cells and batteries, pulls and consumes electricity when needed, but can also push and sell energy when it has a surplus, thus becoming a prosumer. 

This objective can be supported by accurate and dependable forecasting of energy-related data, which is crucial for optimizing decision-making processes at the grid level. Currently, the difference in peak and low demand is significant, resulting in grid inefficiencies. To combat this peak shaving is required~\cite{SHEN2014494}. \gls{dsm} in peak shaving applications is the reduction of energy consumption during peak demand periods~\cite{thakur2016demand}. One approach to \gls{dsm} is \gls{dr}, which encourages end users to shift their use to off-peak hours by increasing tariffs during high demand allowing for a more efficient system utilizing more renewable usage and in turn reducing CO\textsubscript{2} emissions~\cite{CHEN2018125}.

\textbf{Households}
The primary goal of households is to reduce their electricity as much as possible without sacrificing comfort. Occupant behavior significantly influences energy consumption so two households with very similar appliances can have very different energy profiles~\cite{Wang2011}. Research has shown that providing the consumer with individual appliance electricity usage can induce behavior change that can lead to up to 15\% efficiency improvement~\cite{EPRI2009, Darby2006}. Providing such data to the consumer is difficult as sub-metering individual appliances is expensive, one solution to this problem is energy disaggregation, which is the task of inferring individual appliance consumption from the aggregate power signal~\cite{HART1992}. However, due to a lack of a comprehensive multi-region dataset, it is difficult to develop models that could be deployed across different countries and cultures.

To summarize, it can be seen that applications such as electricity flexibility forecasting, peak shaving, demand forecasting, disaggregation, etc. empower the three classes of stakeholders to design better policies and incentive mechanisms in the case of government and regulatory bodies, better and more agile technologies empowering innovative business models in the case of DSOs as well as improved end-user awareness through specific data-driven individualized communication in the case of household consumers. As we focus on domestic electricity consumption, the household is naturally the fundamental unit for data collection, representation, and analysis, and the mentioned applications leverage per household or various types of aggregations of measured data. While there are several per household openly available datasets, their diversity in terms of measured parameters, sampling rate, and duration requires investing extensive effort in pre-processing and harmonization, therefore preventing larger-scale multi-regional data studies for more robust application development. For instance, existing disaggregation models are developed on max 5 datasets with 29 appliances\cite{BERTALANIC2024107318}, and demand forecasting models are usually limited to a single region\cite{AHMAD2020102052}. To mitigate these shortcomings, we provide 1) a multi-region harmonized dataset to serve developing more generic ML models and 2) a structured and enriched form of this data embodied as a knowledge graph that can be semantically queried and is interoperable with other open structured knowledge bases such as Wikidata and DBpedia.
To contribute to better household energy management strategies, this paper proposes the creation of uniform, structured household electricity consumption data compiled from multiple regions globally. Our main contributions are:

\begin{itemize}
\item We provide a multi-region household electricity consumption dataset from various parts of the world compiled from existing publicly available datasets in a uniform format enabling data analysis and \gls{ml} tasks such as energy disaggregation, load demand forecasting, and appliance ON/OFF classification.

\item We propose a comprehensive multi-region electricity \gls{kg} of residential households across the world. The \gls{kg} contains detailed household and appliance metadata and is linked with external \gls{kg}s such as WikiData and DBpedia allowing for extensive analysis and semantic querying.

\item We provide a model training workflow with a pre-trained ensemble of InceptionTime models adapted for appliance ON/OFF multi-label classification. The workflow encompasses 64 appliances from  10 open source datasets across different regions of the world, making it the most comprehensive to date.
\end{itemize}

\subsection*{Related Work}

As discussed in the introduction, disaggregation and load forecasting are the enabling data-driven techniques for reducing CO\textsubscript{2} footprint. To date, existing studies for such techniques employ a limited number of datasets 2-5 usually located on the same continent~\cite{BERTALANIC2024107318,kelly,krystalakos,BONFIGLI20181461} and a limited number of appliances. To our knowledge  the current highest number of datasets used is 5 with 29 appliances~\cite{BERTALANIC2024107318}. A more robust evaluation of these techniques necessitates the utilization of a larger dataset encompassing a broader spectrum of appliances and geographically dispersed locations. Furthermore, only a few studies are entirely replicable by also providing simulation scripts, code, or the developed model itself~\cite{Zhang_Zhong_Wang_Goddard_Sutton_2018, Klemenjak2020}. For this purpose, we provide a multi-region household consumption dataset comprising of 20 datasets in a uniform format along with a data preprocessing pipeline and model training workflow enabling further research and development in this field. Unlike traditional tools like NILMTK~\cite{10.1145/2602044.2602051} which focus on evaluating the accuracy of NILM algorithms, our approach emphasizes knowledge graph generation and the integration of external metadata, enabling analyses for broader insights into energy consumption patterns.

In the domain of energy data and smart home applications, the use of \gls{kg}s has become increasingly prominent. Several key research efforts have significantly contributed to this field. For instance, the study in~\cite{8473447} introduces an innovative method for developing smart home applications by leveraging a runtime \gls{kg} approach, significantly reducing coding requirements by about 85\%. This model conceptualizes smart home scenarios and abstracts smart device manageability into runtime \gls{kg}s, including an automatic generation mechanism for smart home applications to minimize manual coding.

Another notable contribution is from~\cite{10.1007/978-3-030-80418-3_33}, which presents a method for enhancing \gls{ml} forecasters in smart homes through the integration of heterogeneous \gls{iot} and Smart energy data into a \gls{kg}. This research focuses on transforming \gls{iot} data into \gls{rdf} format suitable for \gls{kg}s and investigates the impact of this transformation on prediction algorithms. The methodology maps \gls{iot} data to ontologies and utilizes federated learning, thus enhancing learning capabilities while maintaining data privacy.

Furthermore, ~\cite{FAN2019113395} proposes a novel graph mining-based methodology for analyzing and visualizing building operational data, thereby enhancing building energy management. The authors develop a technique to transform operational data into graphs, analyzed using frequent subgraph mining to identify operational patterns. This approach integrates multi-relational and hierarchical building data, offering valuable insights for energy management.

These studies collectively underscore the versatility of \gls{kg}s in improving energy management and smart home systems. They offer diverse perspectives and methodologies, highlighting the breadth of innovation in this field. Our proposed electricity \gls{kg} addresses a key limitation in existing research by focusing on residential consumption across multiple global regions. The integration of detailed household, location metadata, and appliance-level data enables a level of granular household analysis currently unsupported by other publicly available datasets.

\section*{Methods}

To develop the proposed \gls{kg} we employ a standard development process consisting of the following main steps: identifying data, constructing the \gls{kg} ontology, extracting knowledge, processing knowledge, constructing the \gls{kg}, and maintaining the \gls{kg}~\cite{10.1145/3522586}. This section details the methods used to implement the respective steps and also explains the process of generating the harmonized data.

\subsection*{Existing Open Source Datasets}

\begin{table*}[!h]
	\centering
	\footnotesize

		\begin{tabular}{c|cccccc}
			\toprule
			
			Dataset
			& Sampl. rate 
			& Time
			& Households
			& Appliances
			& Location 
			& Submetered 
			
			\\\midrule
			UK-DALE~\cite{UK-DALE} & 6s 	& 4.3y 	& 5 	& 53 & United Kingdom, Europe & Yes \\
			ECO~\cite{beckel2014nilm}		& 1s 	& 8m 	& 6 	& 21 & Switzerland, Europe & Yes \\
			REFIT~\cite{murray2017electrical} 	& 8s 	& 2y 	& 20  & 23 & United Kingdom, Europe & Yes \\
			DRED~\cite{10.1145/2821650.2821659} 	& 1s 	& 6m	& 1  & 11 & Netherlands, Europe & Yes\\
                SustDataED2 ~\cite{Pereira_2023} 	& 2s 	& 96d	& 1 & 18 & Portugal, Europe & Yes\\
			HEART~\cite{HEART} 	& 1s 	& 1m	& 4  & 13 & Greece, Europe & Yes\\
                DE\_KN~\cite{openpowersystemdata2020} 	& 1min 	& 2.5y	& 11  & / & Germany, Europe & Yes\\
			DEDDIAG~\cite{DEDDIAG_2021} 	& 1s 	& 3.5y	& 15  & 14 & Germany, Europe& Only household 8\\
                IDEAL~\cite{Pullinger2021} 	& 7s 	& 2y	& 255  & / & United Kingdom, Europe & Some households \\
                UCIML~\cite{misc_individual_household_electric_power_consumption_235} 	& 1min 	& 4y	& 1  & / & France , Europe & No\\
                LERTA~\cite{fularz_2021_5608475} 	& 6s 	& 1.5y	& 4  & / & Poland, Europe & No\\
                SustData~\cite{sust1} 	& 1min 	& 3.1y	& 50  & / & Portugal, Europe & No \\
                \midrule
			REDD~\cite{REDD}	& 1s 	& $>$1m & 6		& 17 & Massachusetts, USA, North America & Yes\\
			HES~\cite{HESDataset2023} 	& 7s 	& 5m	& 1  & 39 & Canada, North America & Yes\\
                EEUD~\cite{EEUD} 	& 1min	& 1y	& 23  & 7 & Canada, North America & Yes\\
                HUE~\cite{HUE} 	& 1h 	& 3y	& 22  & / & Canada, North America & No\\
                ECD-UY~\cite{Chavat2022} 	& 15min 	& 1.8y	& 110953  & / & Uruguay, South America & Some households\\
			\midrule
			IAWE~\cite{batra2013s} 	& 1s 	& 73d 	& 1  	& 9	& India, Asia & Yes\\
			ENERTALK~\cite{shin_lee_han_yim_rhee_lee_2019} 	& 1s 	& 4m	& 22  & 7 & South Korea, Asia & Yes\\
                PRECON~\cite{PRECON} 	& 1min 	& 1y 	& 42  	& /	& Pakistan, Asia & No\\
			\bottomrule
		\end{tabular}
        \caption{Summary of the appliance monitoring datasets used in this work.}
        \label{tab:dataset-description}
\end{table*}

In order to develop the \gls{kg}, we first identify openly available datasets that are suitable for the electricity characterization of households and are of sufficient quality. Table~\ref{tab:dataset-description} summarizes the available open-source datasets that are suitable for the proposed \gls{kg}. The first column of the table identifies the dataset by name,
From the second column, it can be seen that the sampling rate of electricity measurements ranges from every second (1s) in most datasets to every hour (1h) in the HUE dataset. Other datasets have intermediate rates like 6s (UK-DALE), 8s (REFIT), and 7s (HES). The sampling rate is particularly important for \gls{ml} applications where datasets typically require a consistent sampling rate that can be achieved through upsampling and downsampling. Additionally, tasks such as disaggregation are not possible with sampling rates of 1 min and above. From the third column, it can be seen that the duration of the measurements varies from one month (HEART) to 4.3 years (UK-DALE). Other datasets cover periods from several months to a few years. The length of the datasets is crucial for analyzing occupant behavior, as a sufficiently large sample size is necessary to conduct meaningful analysis. The fourth column tells us about the number of households present in the dataset this information is important as for example from one household in the DRED dataset it's difficult to draw conclusions about the other households in the area, but if we have a representative sample of the area, for example, 110 953 households in ECD-UY dataset, we can do city or even country level electricity consumption analysis. The number of unique appliances from the fifth column is important because it informs on how complete the dataset is. For example ENERTALK dataset has only 7 unique appliances which means that very likely not all appliances in the household are submetered. The sixth column of the dataset provides information on the countries and continents where the households are located, predominantly in Europe and North America, with three datasets representing Asia. Geographical location is crucial in understanding differences in electricity consumption, as factors such as climate and cultural practices significantly impact electricity usage patterns. The last column tells us if the dataset contains appliance sub-meter data or only the aggregate power consumption data. This information is important so we know which datasets to pick for different \gls{ml} tasks.

\subsection*{Knowledge Graph and Harmonized Data Development Methodology} 

\begin{figure}[!ht]
    \centering
    \includegraphics[width=0.9\linewidth]{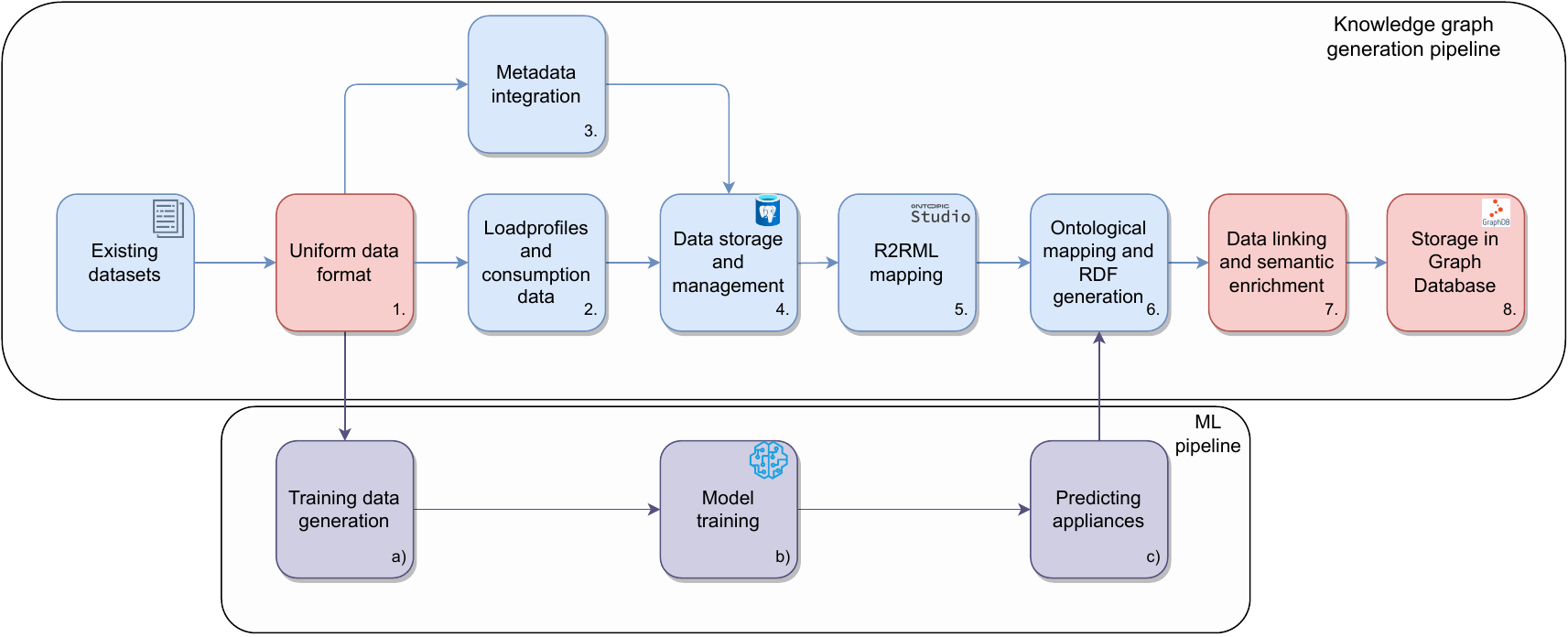}
    \caption{Knowledge graph development methodology.}
    \label{fig:pipeline}
\end{figure}

The proposed \gls{kg} and harmonized data development process is illustrated in Figure~\ref{fig:pipeline} and consists of a \gls{kg} generation pipeline and an \gls{ml} branch. We utilize the \gls{ml} branch to label which appliances are present in the datasets with missing sub-meter data. The data processing pipeline consists of eight data processing steps indicated by blue and red colors in Figure~\ref{fig:pipeline}, namely: 1. Uniform data format, 2. Load profiles and consumption data, 3. Metadata integration, 4. Data storage and management 5. \gls{r2rml} Mapping, 6. Ontological mapping and \gls{rdf} generation, 7. Data linking and semantic enrichment 8. Storage in a graph database. Data contributions proposed in this paper are provided as results of steps 1, 7, and 8 marked with red. The three \gls{ml}-related steps are colored in purple and consist of a) Training data generation, b) Model training, and c) Predicting appliances. The following subsections elaborate on each of the steps while all the code for these steps is available as open source~\footnote{\url{https://github.com/sensorlab/energy-knowledge-graph}}.

\subsubsection*{KG Generation Pipeline}

The \gls{kg} generation pipeline structures electricity usage data into a queryable graph. It starts by standardizing data to a uniform format for integration. Load profiles and consumption metrics are calculated to analyze electricity consumption. Metadata is added for socio-economic and geographical contexts. The data is then stored in a PostgreSQL database and transformed into \gls{rdf} triples using \gls{r2rml} and Ontopic Studio, based on established ontologies. This \gls{rdf} data is linked to external databases like \href{https://www.wikidata.org/} {Wikidata} and \href{https://www.dbpedia.org/}{DBpedia} to enhance its value. Finally, the data is stored in a \href{https://blazegraph.com/}{Blazegraph graph database}, enabling access and querying through a SPARQL endpoint. A SPARQL endpoint provides an interface for querying the database. This endpoint allows users to retrieve specific information and explore relationships within the data using the SPARQL query language. All these steps and their implementation details are described in the following text.

\paragraph{Uniform Data Format:} As discussed in the previous section, the existing data-driven studies related to electricity noticed insufficient measured metrics, sample size, time period, and geographical scale ~\cite{shin_lee_han_yim_rhee_lee_2019}. To create a multi-regional harmonized dataset that can support the stakeholders in their goals by providing a good characterization of the households, we first analyze existing open source datasets as listed in Table ~\ref{tab:dataset-description}, of sufficient quality and sampling frequency that could be easily extended with other datasets generated by regular homes equipped with standard low-frequency smart meters as per European Union and UK technical specifications \footnote{\url{https://www.dlms.com/core-specifications/\#COSEM}}. The datasets originally come in a variety of formats and measurement units; for instance, some are measured in watts while others are in kilowatts. Additionally, the storage of data varies, with some datasets distributed across multiple CSV files per house, others consolidated into one file per house or even multiple houses recorded in a single file. These discrepancies pose challenges in developing preprocessing pipelines without first standardizing the datasets. During the data harmonization phase, we address these issues by creating specific parsers for each dataset to achieve a consistent format across all data. The harmonized datasets can then be utilized further by our open-source tools to structure and enrich them to create the \gls{kg} as depicted in Figure  ~\ref{fig:pipeline}. Following this standardization phase, the data is organized into a nested dictionary structure and provided as pickle files for each dataset. In this structure, each dataset is keyed by the name of the household, leading to a secondary dictionary. This secondary level is composed of keys representing individual appliances, with the corresponding values being the dataframes specific to each appliance containing a datetime index and power consumption in watts. This multi-tiered approach to data organization not only streamlines data storage but also facilitates more efficient and targeted data retrieval for subsequent analyses.

\begin{figure}[ht]
    \centering
    \includegraphics[width=0.9\linewidth]{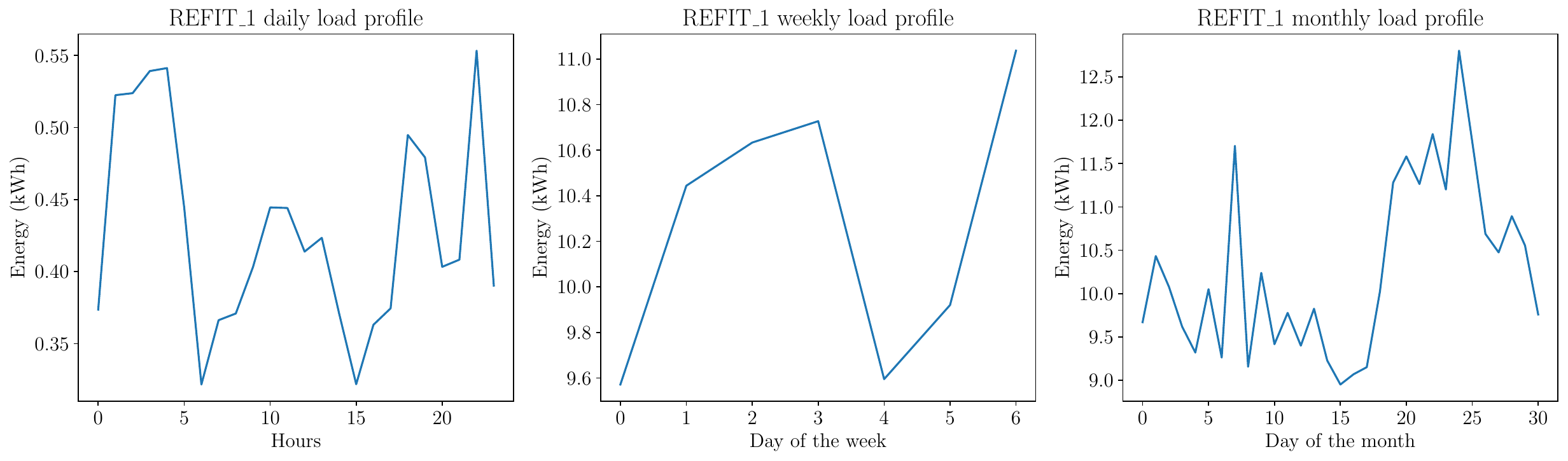}
    \caption{Household load profile example.}
    \label{fig:loadprofile_example}
\end{figure}

\paragraph{Load Profile and Consumption Data Calculation:} Load profiles are computed for each household and each appliance on a daily, weekly, and monthly basis by averaging hourly energy consumption in kWh (for daily profiles) and aggregating data on a daily basis (for weekly and monthly profiles), aiding in the understanding of user behavior and consumption patterns. Figure~\ref{fig:loadprofile_example} illustrates daily, weekly, and monthly load profiles from household 1 in the REFIT~\cite{murray2017electrical} dataset. The y-axis shows energy used (kWh), while the x-axis represents the time of day (daily), or day of week/month (weekly/monthly). Such information offers insights for distributors and policymakers to aid in reducing peak demand~\cite{TROTTA2020115399}. In this step, we also calculate the electricity consumption data at a household and appliance level. We compute the average daily consumption in kWh and additionally, for appliances, we also determine the average ON/OFF event power consumption in kWh. We utilize this data to calculate the carbon footprint of the household, which can help governments shape policies to reduce the carbon footprints of households with high emissions through subsidies and taxes. Additionally, this data is also useful for the residents of the household, who can alter their behavior in an attempt to reduce their carbon footprint by aligning their consumption with times of peak renewable production, if possible.
    
\begin{figure}[!h]
    \centering
    \includegraphics[width=0.65\linewidth]{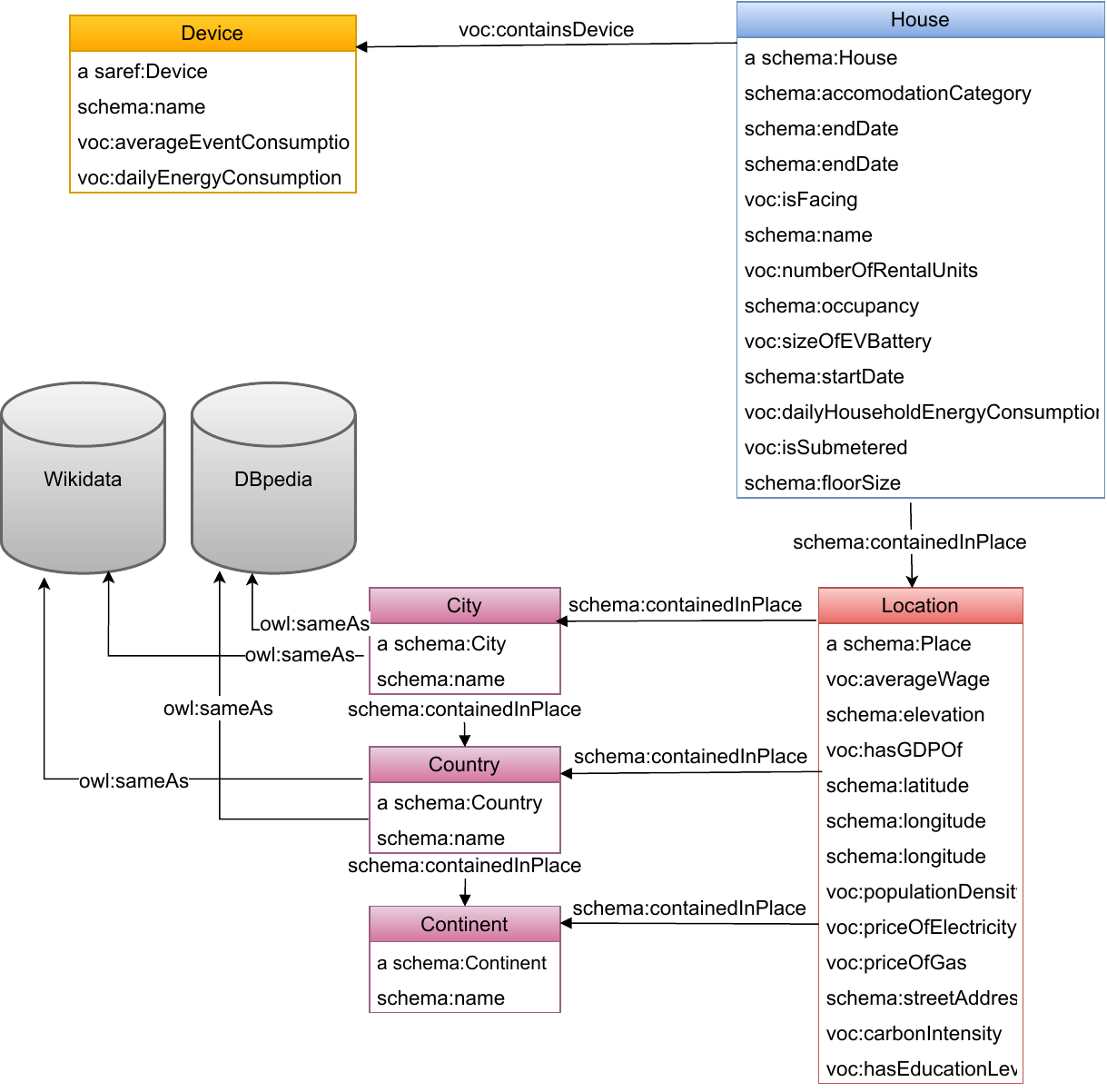}
    \caption{The ontology used for the proposed knowledge graph.}
    \label{fig:data-model}
\end{figure}

\paragraph{Metadata Integration:} The datasets are augmented with spatial and household-related metadata, including geographical coordinates and or city/country information. This data is extracted from the datasets if available. We integrate external data linked through geographical coordinates, covering socio-economic metrics such as \gls{gdp}, average income, education attainment levels, and prices of gas and electricity, along with population density. This can be seen in Figure~\ref{fig:data-model} in the $Location$ class marked in red. Such information can help identify regions with higher CO\textsubscript{2} footprint and can help policymakers create targeted subsidies for such areas to reduce their carbon footprint. Moreover, understanding the interplay between income levels and carbon footprints, as evidenced by the finding that lower-income homes have a disproportionately higher energy usage intensity~\cite{10.1145/3538637.3538801}, underscores the importance of leveraging metadata to design equitable and effective transition strategies towards low-carbon heating solutions.

\paragraph{Data Storage and Management:} Processed, and enriched with metadata, load profiles, and consumption data, the households are organized within a PostgreSQL database to support efficient data management and retrieval. This database also serves as an interim storage solution for our data before we generate triples using an \gls{r2rml} mapping as described in the next paragraph. This method of using relational databases as an intermediary step in \gls{kg} development is a common practice~\cite{10.1145/3522586}. The data regarding households is stored in a $Household$ table, which is linked through foreign keys to a $Location$ table containing location-specific metadata and a $Device$ table detailing appliance information. This configuration also facilitates the use of PostgresML, enabling the execution of \gls{ml} operations directly within the database using SQL, leveraging the stored data.

\paragraph{Ontological Mapping and RDF Generation:} To generate the \gls{rdf} triples used to populate the \gls{kg}, we need a mapping from the PostgreSQL database tables to \gls{rdf} triples. For this, we utilize \gls{r2rml}, a mapping language that maps from relational databases to \gls{rdf} triples. To generate this mapping we utilize Ontopic Studio an \gls{r2rml} no code mapping tool, where we map from the PostgreSQL database tables to \gls{rdf} triples, guided by the SAREF and schema.org ontologies, along with a custom ontology for domain-specific properties like gas and electricity prices. The ontology is visualized in Figure~\ref{fig:data-model}. Taking the household table as an example, we map the \textit{house\_size} column to the predicate \textit{schema:floorSize}. This mapping means that each row in the $house\_size$ column generates a corresponding triple in the \gls{kg}.

\paragraph{Data Linking and Semantic Enrichment:} 
In our research, we aim to significantly enhance the interoperability and value of our \gls{kg} by linking it with established knowledge bases such as Wikidata and DBpedia, focusing particularly on cities and countries. For country linkage, we directly query the SPARQL endpoints using unique country names, enabling us to associate our \textit{schema:Country} entities with their corresponding entities in Wikidata and DBpedia through the \textit{owl:sameAs} predicate. City linkage presents a greater challenge due to the common occurrence of cities sharing names across different locations. To address this, we initiate our process by querying for settlements within a 50-kilometer radius of a household's coordinates from Wikidata and DBpedia endpoints. We then apply string matching using Python's FuzzyWuzzy library to find the correct city from the list of settlements. If an exact match is not found, we select the nearest city to the household coordinates for linkage. Finally, we link our \textit{schema:City} entity to the matching entity in Wikidata or DBpedia using the owl:sameAs predicate, ensuring accurate and meaningful connections within our \gls{kg}. The newly generated triples are then inserted into our \gls{kg} through the SPARQL endpoint.

\paragraph{Storage in Graph Database:} 
\gls{rdf} triples are stored in a Blazegraph\footnote{\url{https://blazegraph.com/}} database, accessible at~\footnote{\url{https://sparqlelec.ijs.si/sparql}}, which supports SPARQL queries and ensures data persistence. Access to a read-only SPARQL endpoint is facilitated via an nginx reverse proxy. The choice of Blazegraph as our data store was influenced by its high performance and open-source nature.

\subsubsection*{Machine Learning Pipeline}
The \gls{ml} pipeline starts by creating a training dataset from sub-metered electricity usage datasets. This involves standardizing appliance names, resampling data to a consistent interval, and correcting data anomalies. The dataset is segmented and balanced to simulate various household energy consumption patterns.

\begin{figure}[!h]
    \centering
    \includegraphics[width=1\linewidth]{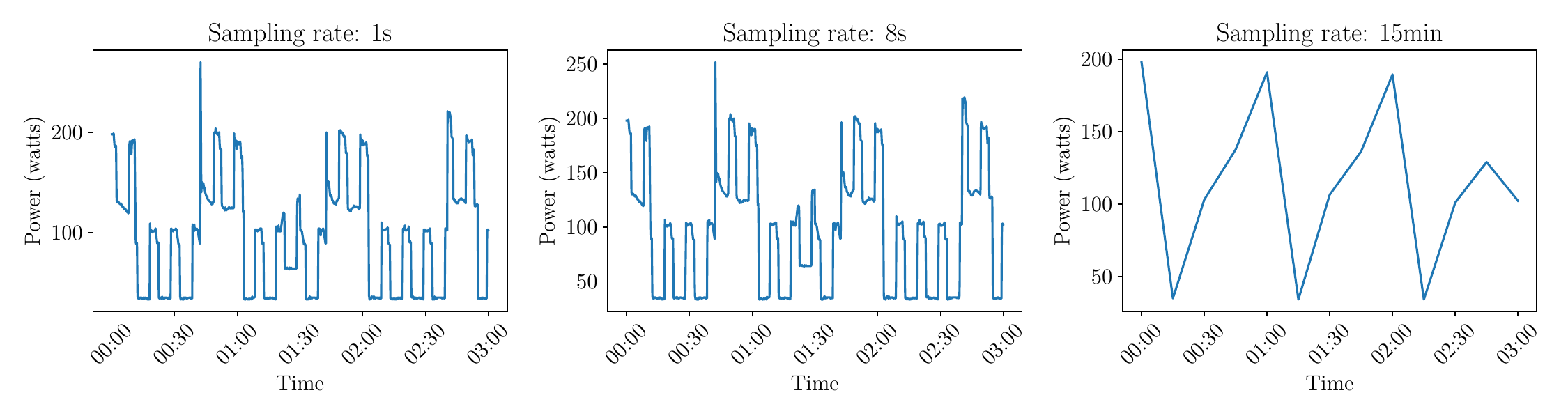}
    \caption{Impact of Sampling Rate on the Visibility of Appliance Usage Events Over Time.}
    \label{fig:sampling_rates}
\end{figure}

\begin{figure}[!h]
    \centering
    \includegraphics[width=0.9\linewidth]{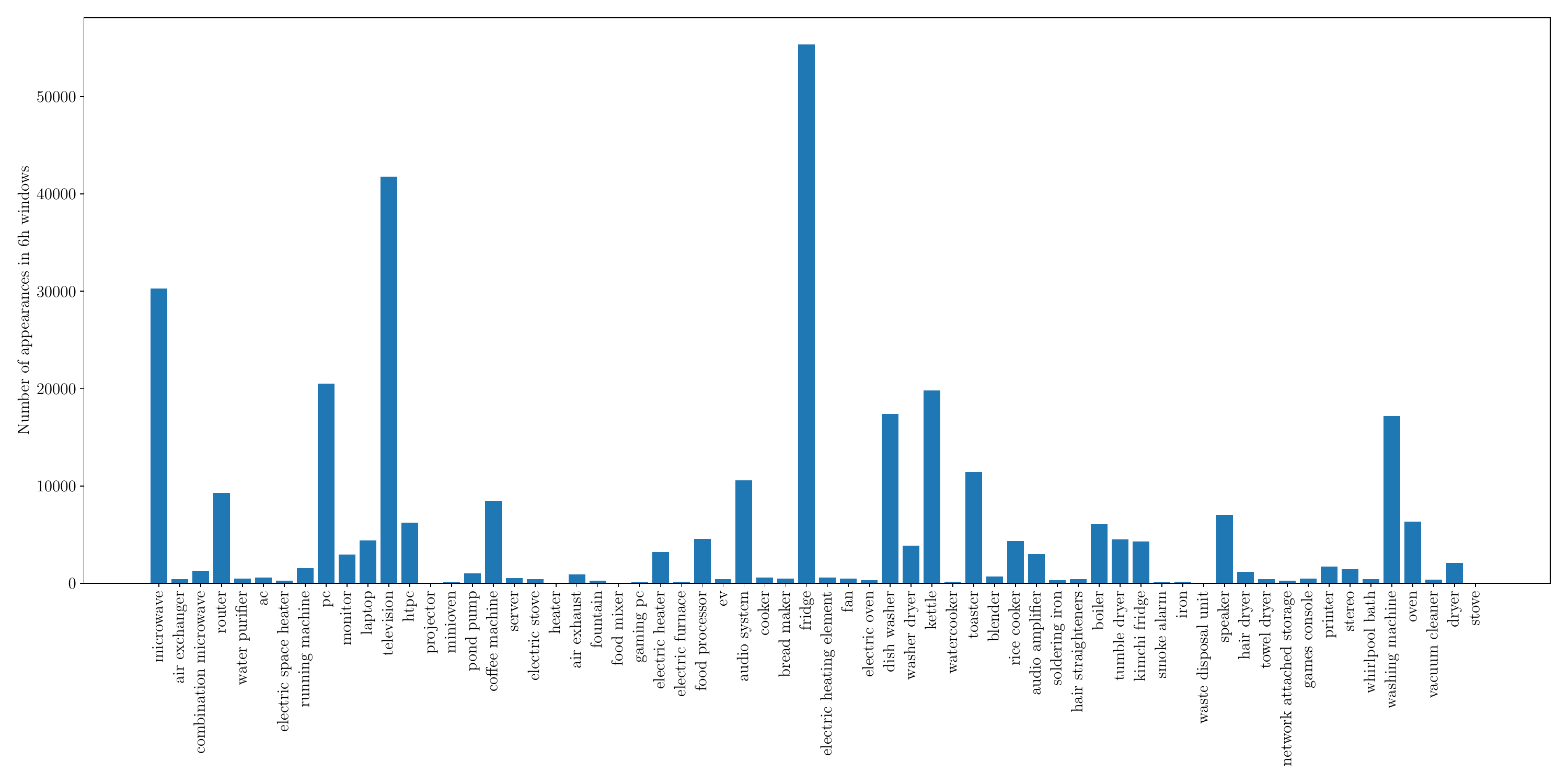}
    \caption{Distribution of appliance representation within data windows. The frequency of each appliance type indicates the number of windows in which its data is present.}
    \label{fig:device_distribution}
\end{figure}

The model, an adapted version of InceptionTime~\cite{IsmailFawaz2020}, is trained on this dataset to predict the appliance's ON/OFF status, employing a multi-label classification approach. Training involves an ensemble of models to ensure robustness, with adjustments for learning rate and early stopping to optimize performance. The pipeline aims to accurately identify appliances present in the household just from the aggregate consumption. These predictions are then added to the \gls{kg} via a SPARQL query.
\paragraph{Training Data Generation:}
 We utilize a comprehensive array of sub-metered datasets with sufficient sampling rates. Datasets with lower sampling rates can not be used for appliance ON/OFF classification as they are not detailed enough to be able to identify individual appliance traces. Figure~\ref{fig:sampling_rates}  highlights the loss of fine-grained patterns essential for appliance ON/OFF classification at the 15-minute sampling which is the standard in the field of energetics. This loss of resolution contrasts with the detail preserved at 1s and 8s sampling rates, which allows for the distinction of individual appliance activity.We use DEDDIAG~\cite{DEDDIAG_2021}, DRED~\cite{10.1145/2821650.2821659}, ECO~\cite{beckel2014nilm}, ENERTALK~\cite{shin_lee_han_yim_rhee_lee_2019}, HEART~\cite{HEART}, HES~\cite{HESDataset2023}, IAWE~\cite{batra2013s}, REDD~\cite{REDD}, REFIT~\cite{murray2017electrical}, and UK-DALE~\cite{UK-DALE}, to train our model. The training process begins with data harmonization, where we employ string matching techniques to unify the nomenclature of appliances across these datasets, such as equating 'fridge' with 'refrigerator'. We exclude appliances, such as outlets due to the ambiguity in what is connected to them. All datasets are then resampled to a uniform 8-second interval, a decision informed by the lowest sampling rate present in the REFIT~\cite{murray2017electrical} dataset, while also adhering to the EU technical specifications dictating that smart meters must have a sampling rate of 10 seconds or less\footnote{\url{https://www.dlms.com/core-specifications/}}. Since power consumption cannot be negative, any negative values encountered in the data are treated as anomalies and reset to zero. Following this, we segment the data from each appliance into 6-hour windows. This window size was chosen to capture the full range of potential appliance operation cycles, including longer-running appliances such as washing machines. We explored various window sizes (30 minutes, 1 hour, 3 hours, and 12 hours), finding that window sizes within this range had a negligible impact on model performance. Windows lacking significant appliance activity, containing too many missing values, or exhibiting unrealistic values are discarded. To mitigate potential bias due to certain appliances dominating the dataset with a higher frequency of representative data windows as seen in Figure~\ref{fig:device_distribution}, We simulate synthetic households for dataset balancing. We sample the number of appliances based on a normal distribution and randomly select the corresponding number of appliances from our entire appliance pool. For each chosen appliance, a random 6-hour window of data from that appliance is selected, and the energy consumption from the selected appliances is aggregated to calculate the total energy consumption measurement. This aggregated data, forming a vector of length 2688 representing household power consumption sampled every 8s, is then normalized using min-max normalization and fed into our ensemble of models. For dataset construction, we generate 100,000 of these windows, allocating 80,000 for model training and 20,000 for testing.

\begin{figure}[!h]
    \centering
    \includegraphics[width=1\linewidth]{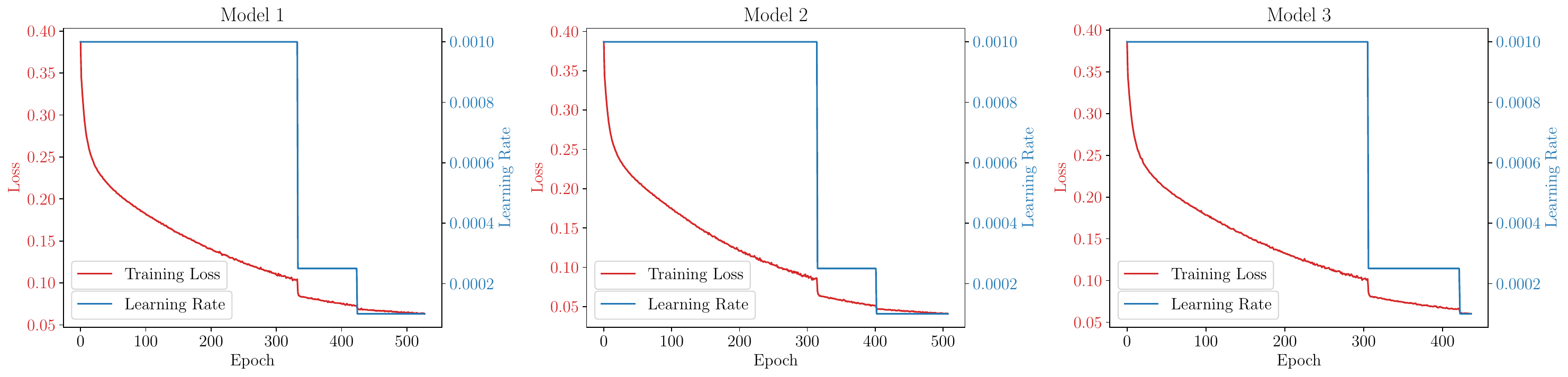}
    \caption{Training Loss and Learning Rate over Epochs.}
    \label{fig:loss_lr}
\end{figure}

\begin{figure}[!h]
    \centering
    \includegraphics[width=0.5\linewidth]{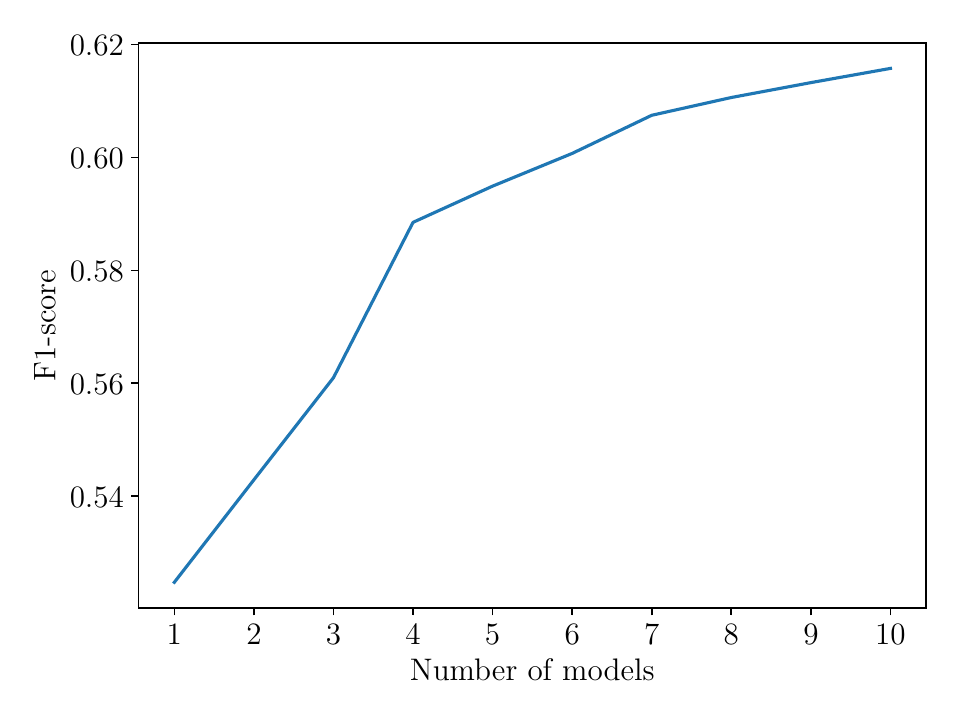}
    \caption{F1-score vs Ensemble size.}
    \label{fig:ens_size}
\end{figure}

 \paragraph{Model Training:}
To develop an appliance classifier using the submetered datasets from Table~\ref{tab:dataset-description} and the training data generation from step a, we selected the InceptionTime~\cite{IsmailFawaz2020} architecture, a state-of-the-art time-series processing neural network proven to have excellent performance on multi-class timeseries classification which we adapted for multilabel-classification. This adaptation involves modifying the loss function to Binary Cross-Entropy and substituting the final layer's activation function with a sigmoid to accommodate for multi-label classification. The architecture consists of an ensemble of Inception classifiers and the ensemble size as well as the input size can be tuned to accommodate different tasks. During the model training step, we used an ensemble of 10 models with an input size of 2688. We opted for an ensemble of 10 models because it enhances the F1-score by over 10\% compared to using a single model. We did not increase the ensemble size beyond this as the performance gains show diminishing returns; the improvement in the F1-score between using 7 and 10 models is less than 1\% at a significant computational cost. This trend is illustrated in Figure~\ref{fig:ens_size}, which plots the F1-score on the y-axis against the number of models in the ensemble on the x-axis. The input represents the aggregate consumption of the household in a 6-hour window with a sampling rate of 8s, and the training target is a binary vector of length 64 that represents the ON/OFF status of appliances in the given time window. The models are trained for up to 1200 epochs, incorporating a learning rate that diminishes when a plateau in performance is detected. Additionally, we implement an early stopping mechanism that halts training upon the models' convergence, which typically occurs around the 450-epoch mark. This training strategy is visualized in Figure~\ref{fig:loss_lr}, which displays the progression of training loss and the adjustments in learning rate for three models from our ensemble. A notable observation from the figure is the initial learning rate reduction, which occurs approximately at the 320-epoch juncture, illustrating the adaptive nature of our training methodology to optimize model performance. With this training methodology, our model achieved a sample average F1-score of 0.58 on the test set with 64 appliances.
 
\paragraph{Predicting Appliances:} 
To apply our trained model to an unlabeled dataset with the aim of detecting the presence of appliances, we follow these steps: First, we standardize an unlabeled, previously unseen dataset, with a sampling rate of at least 8 seconds. If the dataset has a higher sampling rate we down-sample it to 8s. Next, we divide the dataset into segments spanning 6 hours each, discarding any segment without active appliance readings or with excessive missing data. Subsequently, we normalize the data within each segment using a min-max normalization. Finally, these prepared segments are fed into our model. We then average the output probabilities over all the windows. To determine the presence of appliances, we apply thresholding. In this process, a threshold value (in our case, 0.3), chosen based on experiments conducted on labeled data is selected. Any appliance with an average probability exceeding this threshold is considered present within the household.

\section*{Data Records}

\subsection*{Raw Data Dump}
The datasets utilized in our study, along with necessary metadata, are available for download as tar.gz compressed files in our \href{https://github.com/sensorlab/energy-knowledge-graph}{Github} repository. To accommodate varying computational resources, we offer two versions of the data dump: a smaller, 10.4 GB version for processing on less powerful machines, and a complete, comprehensive dump that includes all datasets and households, which is approximately 91.2 GB gigabytes compressed with gzip.
\subsection*{Harmonized Data}
\begin{table}[ht]
\centering

\begin{tabular}{r|rrr}

\toprule
    & Europe & Americas & Asia \\ \midrule
Total household-level aggregated power consumption meas. & 1,085,720,074      & 5,643,637,499       & 169,925,751    \\
Total individual appliance power consumption meas.  & 2,647,433,215      & 99,730,950      & 473,770,708    \\
Mean household-level aggr. power consumption meas. per household & 3,102,057       & 50,839       & 2,614,242    \\
Mean individual appliance power consumption meas. per household & 7,564,094      & 898      & 7,288,780 \\ 
Total duration of household-level aggr. power consumption in years & 307 & 164,321 & 42 \\
Total duration of individual appliance consumption in years & 623 & 83 & 0.25 \\
Total number of appliances & 585 & 139 & 81 \\
Total number of households & 350 & 111,009 & 65 \\
\bottomrule
\end{tabular}
\caption{Comparison of household-level aggregate and individual appliance power consumption data across continents.}
\label{tab:harmonized_data}
\end{table}

The harmonized dataset derived from the datasets summarized in Table~\ref{tab:dataset-description}  encompasses power consumption measurements from 111,424 households across 20 datasets, converted in a uniform format for further preprocessing and analysis. The datasets contain 6,899,283,324 household-level aggregated power consumption measurements and 3,220,934,873 individual appliance power consumption measurements. A more fine-grained breakdown of the average number of measurements per household, appliance, and region is available in Table~\ref{tab:harmonized_data}. 
The table reveals that the majority of appliance-specific (sub-meter) data is collected from European datasets, with approximately 2.6 billion appliance power measurements, compared to 100 million in the Americas and 473 million in Asia. Despite this, the bulk of the households surveyed are in the Americas, primarily due to the ECD-UY dataset from Uruguay, which includes 110,953 households. This contributes to a higher total of aggregate household power consumption measurements in the Americas. However, when examining the mean aggregate household power consumption measurements per household, the Americas have significantly fewer data points, approximately 50,839, compared to 3 million in Europe and 2.6 million in Asia. This notable disparity can largely be attributed to the difference in sampling rates; the ECD-UY dataset from Uruguay uses a 15-minute sampling interval, whereas most European and Asian datasets feature more frequent sampling rates, often less than one minute. This disparity is also reflected in the duration and scale of data collection; despite having only 350 households spanning 307 years of aggregate household power consumption data, compared to the 164,321 years of aggregate household power consumption data from 111,009 households in the Americas. Yet, Europe has only five times fewer aggregate household power consumption measurements than the Americas, further highlighting the more frequent data sampling in currently available European datasets. This harmonized dataset can be generated by parsing the raw data dump using our parsers or downloaded from our \href{https://github.com/sensorlab/energy-knowledge-graph}{Github} repository.

\subsection*{Knowledge Graph}

Our \gls{kg} is composed of 791,813 nodes (i.e instances of subjects or predicates) interconnected through 1,577,483 instances of predicates totaling 1,577,483 triples. It is generated from 6 unique subject/object concepts and 38 unique predicates that form the vocabularies summarized in Table~\ref{tab:knowledge_graph_summary}. It encompasses 111,423 instances classified under \textit{schema:House} class representing individual households, and details such as average daily consumption and load profiles on 113,359 unique submeters and meters that measure per appliance or aggregated consumption. The households span 54 unique locations across 14 countries and 12 cities, underscoring the \gls{kg}'s comprehensive global perspective and its potential utility for diverse regional energy analyses and applications. Each location instance contains various metadata pertaining to the location, such as GDP, average wage, electricity prices, etc., and is also linked to external knowledge bases such as WikiData and DBpedia further expanding the available location-specific properties. Access to this \gls{kg} is facilitated through a \href{https://sparqlelec.ijs.si/sparql}{SPARQL endpoint}, and for those interested in a more granular exploration, we offer a \href{https://github.com/sensorlab/energy-knowledge-graph}{downloadable} dump of the \gls{rdf} triples constituting the graph. We also provide a visual representation of our \gls{rdf} data with \href{https://elkg.ijs.si/resource/public-households/REFIT_1}{LodView}, where we display load profile plots and other properties of the \gls{kg}.

\begin{table}[!ht]
\centering
\begin{tabular}{|r|r|}
\hline
\textbf{KG entites} & \textbf{Quantity} \\
\hline
 Total triples & 1,577,483 \\
Total instances of predicates & 1,577,483\\
Total unique predicates & 38 \\
Total instances of nodes/ subject or object  & 791,813 \\
Total unique concepts/ subject or object & 6 \\
\hline
Instances of \textit{schema:House} & 111,423 \\
Instances of \textit{saref:Device} (submeters and meters)  & 113,359 \\
Instances of \textit{saref:Device} corresponding to appliances & 1936 \\ 
\hline
Instances of \textit{schema:Continent} & 4 \\
\hline
Instances of \textit{schema:Country} & 14\\
Instances of DBpedia countries & 14 \\
Instances of Wikidata countries & 14 \\
\hline
Instances of \textit{schema:City} & 12 \\
Instances of DBpedia cities & 12 \\
Instances of Wikidata cities & 12 \\
\hline
\end{tabular}
\caption{Summary of the Knowledge Graph Data.}
\label{tab:knowledge_graph_summary}
\end{table}

\section*{Technical Validation}
\subsection*{Harmonized Data Validation}
In the process of harmonizing data, we analyze open-source datasets, as referenced in Table~\ref{tab:dataset-description}. During this analysis, we eliminate any instances of missing data and convert timestamps to datetime objects, which are then used as indices in pandas. The processed data is organized into a nested dictionary format and preserved as a pickle file. It's important to note that we do not employ interpolation or other methods to fill in missing data, ensuring the integrity and quality of the original dataset remain intact. 

\subsection*{Metadata Validation}
For household-specific metadata, we directly use the information provided in the datasets without modification. In terms of spatial data, we utilize the available household coordinates or, in their absence, the country information. It's important to note that when only country information is available, without precise household locations, we are unable to enrich the dataset with specific details such as elevation and population density. This spatial data then serves as a basis to link with external sources for further enrichment with socio-economic factors as described in more detail in the methodology section.

\subsection*{Mapping Validation}
Using Ontopic Studio~\cite{Calvanese2017}, we create \gls{r2rml} mappings to facilitate the transformation of data from a PostgreSQL database into \gls{rdf} triples. These mappings strictly follow the standards set by schema.org, SAREF, and our custom ontology. The mapping process guarantees error-free conversion of database data into \gls{rdf} triples, thus preserving the data's integrity and ensuring its accurate representation in \gls{rdf} format.

\subsection*{Linking Validation}
We use a custom Python script to link our \gls{kg} to WikiData and DBPedia. We link our \textit{schema:City} and \textit{schema:Country} entities using the \textit{owl:sameAs} predicate to their corresponding counterparts in WikiData and DBPedia. For countries, direct queries to the SPARQL endpoints using country names are sufficient, given the uniqueness of country names. City connections, however, require a two-step process: initially, we query for all settlements within a 50km radius of a household's coordinates. Subsequently, we employ string matching to identify the accurate city. In cases where a direct match proves elusive, we default to the nearest settlement to the household's coordinates, ensuring the most accurate linkage possible.

\subsection*{Predicted Appliances Validation}
We employ a model to categorize the unlabeled datasets, achieving a sample average F1-score of 0.58 across 64 classes. Due to this, the identification of appliances may not always be accurate. Consequently, each household is assigned a property, \textit{voc:isSubmetered}, which indicates the reliability of the appliance data. If true, it signifies that the dataset includes submeter data, confirming the appliances present as ground truth. However, if false, it implies that the appliances were identified using our model, and their accuracy may be uncertain.

\section*{Usage Notes}
\subsection*{Preprocessing and Training Pipeline}
We provide Python scripts for our preprocessing and model training pipeline in our \href{https://github.com/sensorlab/energy-knowledge-graph}{github repository} with detailed instructions on how to use them in the form of README files.

\subsection*{Knowledge Graph}
Households and their metadata can be accessed via the provided \href{https://sparqlelec.ijs.si/sparql}{SPARQL endpoint}, here we provide an example SPARQL query in Listing~\ref{lst:sample_query}, there are more SPARQL query examples available in our \href{https://github.com/sensorlab/energy-knowledge-graph}{github repository}
\begin{lstlisting}[caption={SPARQL query to retrieve all households located in London.}, label=lst:sample_query]
PREFIX schema: <https://schema.org/>
PREFIX data: <http://mydata.example.org/>
PREFIX rdf: <http://www.w3.org/1999/02/22-rdf-syntax-ns#>
SELECT DISTINCT ?housename ?cityName 
WHERE {
?place rdf:type schema:Place .
?house rdf:type schema:House .
?place schema:containedInPlace ?City .
?house schema:containedInPlace ?place .
?house schema:name ?housename .
?City rdf:type schema:City .
?City schema:name ?cityName .
    FILTER(?cityName = "London").
} 
\end{lstlisting}

\section*{Code Availability}
Our source code to preprocess the data and generate the \gls{kg} is available at our \href{https://github.com/sensorlab/energy-knowledge-graph}{Github} repository. Our code is provided under the BSD 3-Clause License.

\bibliography{main}

\section*{Acknowledgements} 

This work was funded in part by the Slovenian Research Agency under the grant P2-0016 and L2-50053. The authors would also like to acknowledge Dr. Gregor Cerar for his very early code contributions. We would like to thank the team at Ontopic Studio for providing a software license to develop the \gls{kg}.

\section*{Author Contributions Statement}
Conceptualization: C.F. and B.B.; Data curation: V.H.; Formal analysis: V.H.; Funding acquisition: C.F.; Investigation: V.H. and B.B.; Methodology: V.H.; Project administration: C.F. and B.B.; Resources: C.F. and B.B.; Software: V.H.; Supervision: C.F. and B.B.; Validation: C.F., V.H. and B.B.; Visualization: V.H.; Writing – original draft: V.H and C.F.; Writing - review \& editing: C.F., V.H. and B.B.
\section*{Competing Interests}
The authors declare that they have no known competing financial interests or personal relationships that could have appeared to influence the work reported in this paper.

\end{document}